\documentclass[doublecolumn]{IEEEtran}
\usepackage{amsmath,amsthm,amsfonts,amssymb,bm,mathrsfs}
\usepackage{subfigure}
\usepackage{graphicx,multicol}
\usepackage{algorithm}
\usepackage{algorithmic}
\usepackage{multirow,booktabs,threeparttable}
\usepackage{color,cite,url}
\usepackage{booktabs}
\usepackage{amssymb}
\usepackage{caption}
\usepackage{tabularx}
\usepackage{makecell}

\newcolumntype{Y}{>{\centering\arraybackslash}X}
\newcolumntype{L}{>{\raggedright\arraybackslash}X}

\graphicspath{{fig/}}

\usepackage{setspace}
\theoremstyle{remark}

\theoremstyle{plain}

\begin{document}

\title{Adaptive Layer Splitting for Wireless LLM Inference in Edge Computing: A Model-Based Reinforcement Learning Approach}

 \author{
 	Yuxuan Chen, Rongpeng Li, Xiaoxue Yu, Zhifeng Zhao, and Honggang Zhang

 	\thanks{Y. Chen, R. Li, X. Yu and H. Zhang are with Zhejiang University, Hangzhou 310027, China, (email: \{cyx00, lirongpeng, sdwhyxx, honggangzhang\}@zju.edu.cn).}

	\thanks{Z. Zhao is with Zhejaing Lab, Hangzhou 310012, China as well as Zhejiang University, Hangzhou 310027, China (email: zhaozf@zhejianglab.com).}
 }

\maketitle
\begin{abstract}

    Optimizing the deployment of large language models (LLMs) in edge computing environments is critical for enhancing privacy and computational efficiency. Toward efficient wireless LLM inference in edge computing, this study comprehensively analyzes the impact of different splitting points in mainstream open-source LLMs. On this basis, this study introduces a framework taking inspiration from model-based reinforcement learning (MBRL) to determine the optimal splitting point across the edge and user equipment (UE). By incorporating a reward surrogate model, our approach significantly reduces the computational cost of frequent performance evaluations. Extensive simulations demonstrate that this method effectively balances inference performance and computational load under varying network conditions, providing a robust solution for LLM deployment in decentralized settings.
    
\end{abstract}

\section{Introduction}
The field of natural language processing (NLP) has recently experienced transformative changes, driven by the rapid advancement of large language models (LLMs) such as GPT-4 \cite{openai2024gpt4} and Gemini \cite{team2023gemini}. These models are highly proficient at generating human-like text \cite{brown2020language, bai2022training, touvron2023llama, wei2021finetuned, le2022bloom}, catalyzing progress across various domains \cite{nijkamp2022codegen, webb2023emergent, jin2024health, roziere2023code}.~
Although LLMs perform well in centralized cloud environments, they face significant scalability and privacy issues \cite{thirunavukarasu2023large, wu2023bloomberggpt} in sensitive applications like healthcare and finance \cite{kaddour2023challenges}, which have driven the exploration of edge computing \cite{mao2017survey, mach2017mobile} as a complementary paradigm. 
Notably, edge computing processes sensitive information locally rather than traversing through a centralized cloud \cite{li2019edge, letaief2021edge}.~Consequently, it minimizes the exposure to potential privacy breaches and unauthorized access. Moreover, edge computing allows for a flexible and distributed architecture, and can accommodate allocated computational resources to specific requirements \cite{abbas2017mobile, pham2020survey}.~Therefore, the integration of edge computing and LLMs empowers LLMs with enhanced personalized and domain-specific generative capabilities \cite{chen2024netgpt}.~However, LLMs' substantial computational demands often exceed the processing capacities of communication-limited user equipment (UE) in radio access network or Internet of Things (IoT) systems \cite{lin2024pushing,patil2024review, 10.36227/techrxiv.23589741.v1}. Correspondingly, split learning and inference \cite{lin2024split, lee2023wireless, qiao2023timely} are proposed to jointly leverage the computing capability of UE and edge nodes (e.g., base stations [BSs]).  

Though most of the existing works \cite{chen2021distributed, ryu2022study, lin2024split,lan2021progressive, karjee2022split, lee2023wireless, wang2023split} focus on careful model splitting to balance the computational and communication costs, splitting different layers of LLMs is quite unique, as the intermediate outputs have consistent dimensions, leading to the same communication cost. Additionally, transmitting tensors between LLM layers over potentially noisy channels could hinder LLM inference performance. As validated in this work lately, given different splitting points, the possible loss induced by unreliable wireless channels produces a significantly diverse impact on model performance. 
These attributes necessitate a shift in focus from optimizing transmission efficiency to managing the computational burden on UE. In LLM deployment scenarios, where the scale and complexity of the models impose significant demands on UE's limited computational resources, the challenge lies in balancing the computational load without degrading inference performance.
Thus, it is critical to identify the optimal splitting point for model inference while combating wireless channel volatility. Such a viewpoint transcends traditional single-step optimization techniques and warrants a new framework accommodating the sequential nature of decision-making.

\begin{figure}[t]
	\centering
	\includegraphics[width=0.495\textwidth]{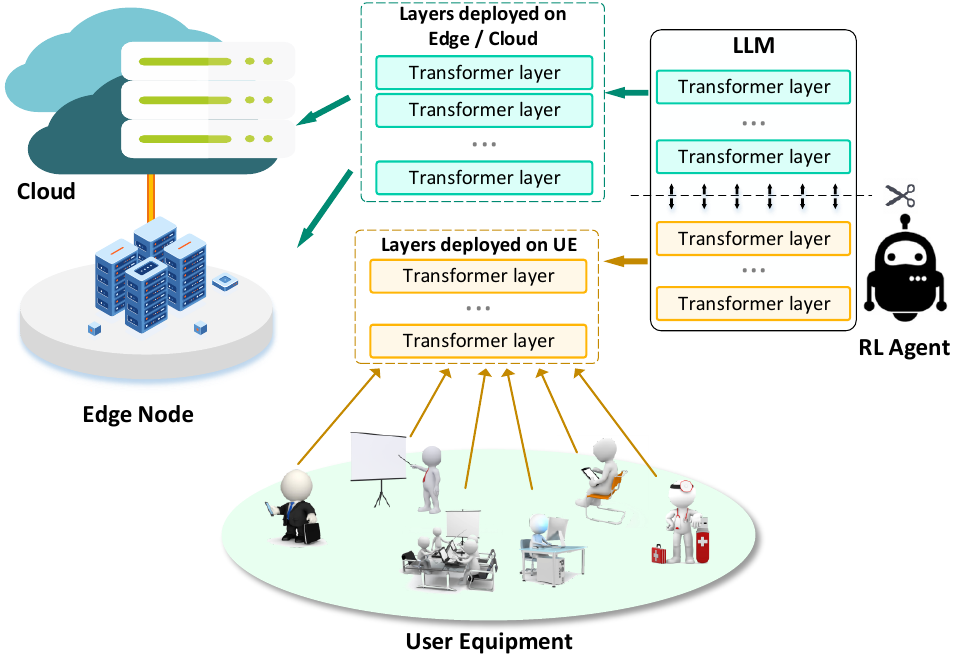}
	\caption{A high-level architecture of the framework, depicting the distribution of the LLM across edge and UEs, highlights the role of the RL agent in managing interactions between the LLM and wireless networks.}
	\label{fig:overview}
\end{figure}

Reinforcement learning (RL) emerges as an apt methodology, known for its proficiency in sequential decision-making tasks \cite{mnih2015human, li2017deep, luong2019applications, qian2019survey}. RL learns optimal strategies over successive iterations, continuously adapting to the dynamic and uncertain edge environment.~The iterative nature of RL, requiring multiple interactions with the LLM to ascertain the rewards of different actions, complements the continuous and unpredictable variations in wireless channels.~Nevertheless, it inevitably incurs significant interaction costs before collecting sufficient records. Fortunately, with the merits in sampling efficiency, model-based reinforcement learning (MBRL) is particularly suited to this context \cite{deisenroth2011pilco, kaiser2019model, shlezinger2021model, moerland2023model}. In particular, MBRL capably simulates the uncertain environment and assesses the potential outcomes of actions, thus enabling a more informed and anticipatory optimization strategy for the dynamical splitting point determination process.

\begin{table*}[tbp]
\centering
\caption{Summary and comparison of related works}
\label{tab:related_work}
\begin{tabularx}{\textwidth}{|l|X|X|}
\toprule
\textbf{Refs} & \textbf{Brief description} & \textbf{Limitations} \\ \midrule
\cite{lin2024pushing} & Discusses the deployment of LLMs at 6G edges, advocating for edge computing to optimize LLM deployment. & Lacks detailed methodologies for handling computational constraints at edge nodes. \\ \hline
\cite{gupta2018distributed} & Introduces split learning, which splits models between clients and servers to avoid transferring raw data. & Focuses on DNNs without considering the unique challenges of wireless channels and high deployment costs associated with LLMs. \\ \hline
\cite{lee2023wireless} & Explores split inference in wireless networks, distributing DNNs for collaborative inference. & Does not account for the specific challenges of LLMs, such as high computational requirements and the impact of wireless channel volatility on performance. \\ \hline
\cite{shlezinger2021model} & Investigates model-based machine learning for communication systems, emphasizing the advantages of predictive modeling in dynamic environments. & Limited focus on LLMs and their unique characteristics, including high processing demands and sensitivity to channel noise. \\ \bottomrule
\end{tabularx}
\end{table*}

As illustrated in Fig.~\ref{fig:overview}, we propose an adaptive layer splitting algorithm for efficient LLM inference in edge computing, where only a few selected transformer layers are provisionally activated at the UE.
Meanwhile, faced with wireless channel fluctuations, we leverage a sample-efficient MBRL-inspired approach to determine the suitable splitting point.
While highlighting the key differences with existing works in  Table \ref{tab:related_work}, the primary contributions of this paper are summarized as follows.
\begin{itemize}
    \item We comprehensively evaluate the impact of LLM splitting points on LLM inference performance under varying channel conditions, and formulate the determination of appropriate LLM splitting points as a sequential decision-making process.
    \item We leverage proximal policy optimization (PPO) \cite{schulman2017proximal} for adaptive splitting point determination and devise a sample-efficient reward surrogate model to facilitate the learning.
    \item We conduct extensive empirical studies to evaluate the robustness and effectiveness of the MBRL-inspired splitting point determination method.
\end{itemize}

The rest of this paper is organized as follows: Section \ref{sec:related} reviews related works, setting the context for our research. Section \ref{sec:model} first describes our system model and highlights the impact of splitting points on LLM inference performance. Afterward, Section \ref{sec:model} presents the formulated problem, while Section \ref{sec:rl} explores an MBRL-inspired splitting point determination solution. Section \ref{sec:simulation} presents our experimental setup and results. Finally, we conclude the paper with future research directions in Section \ref{sec:conclusion}.

\section{Related Works}
\label{sec:related}
\subsection{Edge-Enhanced LLM Deployment}

In the realm of LLMs, \cite{lin2024infinite, chen2024position} provide a comprehensive overview of the challenges faced by LLMs in cloud-based settings, particularly emphasizing the constraints related to data privacy and processing efficiency. 
\cite{satyanarayanan2009case} proposes the concept of edge computing as a viable solution. Subsequently, \cite{chen2024netgpt} proposes an LLM cloud-edge collaboration framework, which utilizes small-scale models deployed at edge nodes to enhance the generative capabilities of cloud-based LLMs. Additionally, \cite{lin2024pushing} emphasizes the importance of optimizing LLM deployment at 6G edges but does not provide specific methodologies for leveraging edge computing capabilities. \cite{dong2024creating} proposes to use LLMs as offline compilers to meet low-latency requirements in edge computing. However, these approaches do not fully address the computational constraints of edge nodes. Our research bridges this gap by introducing a dynamical layer splitting framework to alleviate these limitations.

\subsection{Split Inference in Distributed Computing}

\cite{gupta2018distributed} introduce to split models between clients and servers to avoid transferring raw data, offering new perspectives on distributed computing and privacy-preserving AI. 
\cite{chen2021distributed, ryu2022study, lin2024split} broaden the application of split learning to encompass the distributed deployment of deep neural networks (DNNs) in wireless networks. \cite{lan2021progressive, karjee2022split, lee2023wireless, wang2023split} distribute different portions of DNNs between edge nodes and cloud for collaborative inference task execution, thus reducing response latency and improving the scalability. 

Different from these existing works \cite{lan2021progressive, karjee2022split, lee2023wireless, wang2023split},  LLM splitting presents unique challenges. Due to the large computational demands of LLMs, deploying different layers across UE and edge nodes requires careful consideration of the computational load on the UE. The significant disparity in computational capabilities between heterogeneous devices further complicates the deployment process \cite{zhang2024edgeshardefficientllminference, Ong:EECS-2024-108}. Also, splitting at different points within the LLM significantly impacts the overall inference performance, as demonstrated by our experiments in Section 3. 
To our best knowledge, this belongs to the first efforts to address this important issue, and lays the very foundation for further adaptive layer splitting to balance LLM inference performance and UE computation cost.

\subsection{Reinforcement Learning in Network Optimization}

RL emerges as a pivotal tool for optimizing decision-making processes in dynamic and uncertain environments, as highlighted by \cite{zhu2022alleviating, icarte2023learning}. \cite{yang2023offline, ke2023applying} demonstrate RL's efficacy in enhancing performance optimization within distributed networks, highlighting its potential to adapt and respond to evolving environmental conditions. In the context of wireless network environments, \cite{liu2020optimizing, li2022federated} leverage RL to address the challenges of resource allocation and network traffic management, showcasing its capability to optimize system performance amidst the fluctuating nature of wireless communications.
With its predictive modeling capabilities and remarkable sampling efficiency, MBRL is adept at navigating environments with variable factors \cite{kaiser2019model, egorov2022scalable}. To address computational constraints and wireless channel volatility, our research draws inspiration from MBRL and develops a computation-efficient reward surrogate model to optimize LLM deployment at the edge.

\section{System Model and Problem Formulation}
\label{sec:model}
In this section, we begin with a comprehensive description of the system model, which highlights the deployment of a splitting LLM across wireless network. Afterward, we discuss the impact of the layer splitting point on LLM performance under various channel conditions. Finally, we formulate the channel-aware splitting point optimization problem to balance the UE computational load and LLM inference performance. 

Beforehand, we summarize the mainly used notations in Table \ref{tab:Notations}.

\begin{table}[thb]
\centering
\caption{Notations used in the paper.} \label{tab:Notations}
\renewcommand{\arraystretch}{1.15}
\setlength{\tabcolsep}{2pt}
\begin{tabular}{p{2cm} p{5cm}}
    \toprule[0.75pt]
    \textbf{Notation} & \textbf{Definition} \\
    \midrule[0.5pt]
    $L$ & Total number of layers in the LLM \\
    $p$ & Adjustable splitting point, indicating the number of layers deployed at the UE \\
    $L_U(\cdot), L_C(\cdot)$ & Layers deployed on UE and edge respectively \\
    $x \in \mathbb{R}^{d_{\text{in}}}$ & Input data \\
    $y, \hat{y} \in \mathbb{R}^{d_{\text{mid}}}$ & Intermediate tensor before and after the channel \\
    $\theta_{\text{UE}}, \theta_{\text{edge}}$ & Parameters of the LLM layers deployed on the UE and edge \\
    $h$ & Nakagami-$m$ fading channel gain \\
    $m$ & Nakagami-$m$ fading channel shape parameter \\
    $\Omega$ & Nakagami-$m$ fading channel spread parameter \\
    $n \sim \mathcal{N}(0, \sigma^2)$ & Gaussian distributed noise with mean $0$ and variance $\sigma^2$ \\
    $h_{\text{th}}$ & Threshold for channel gain below which packet loss occurs \\
    $P$ & Probability of packet loss \\
    $z \in \mathbb{R}^{d_{\text{out}}}$ & Inference output provided by the edge \\
    $\theta_{\text{surr}}$ & Parameters of the reward surrogate model \\
    \bottomrule[0.75pt]
\end{tabular}
\end{table}

\subsection{System Model}
\begin{figure*}[t]
	\centering
	\includegraphics[width=0.95\textwidth]{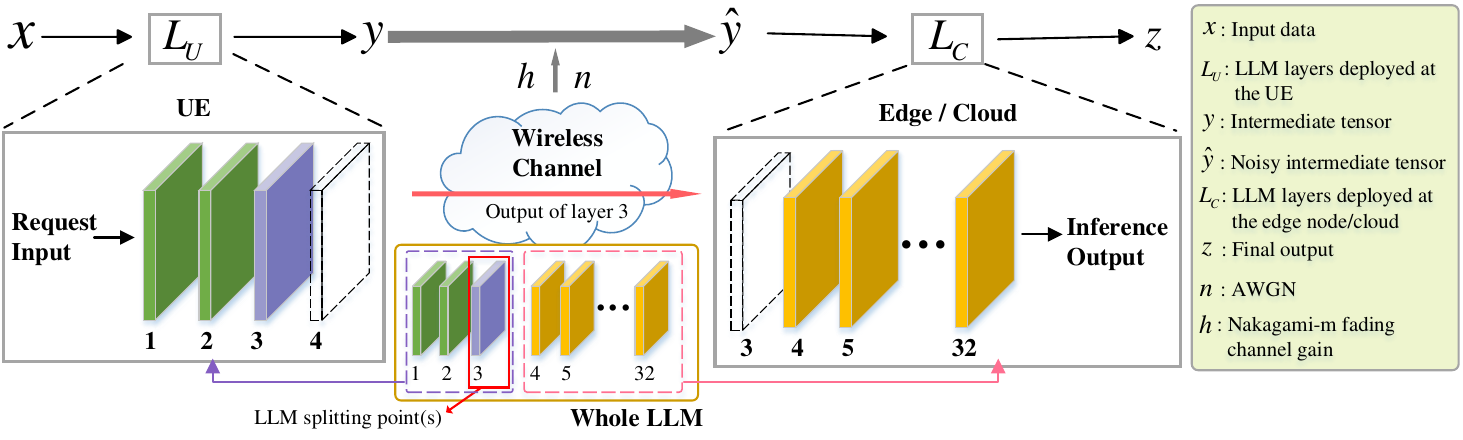}
	\caption{Overview of the split model architecture in wireless channel, with layer 3 designated as the example splitting point. We use the 32-layer LLaMA2-7B model as an example.}
	\label{fig:system_model}
\end{figure*}
To characterize the LLM provisioning with model splitting, we primarily consider a system model illustrated in Fig. \ref{fig:system_model}. Without loss of generality, we assume that for an $L$-layer LLM, the first $p$ layers $L_U(\cdot)$ are deployed at the UE, and the remaining $L-p$ layers $L_C(\cdot)$ are in the edge, where the splitting point $p$ is adjustable. Based on this, for an input $x\in \mathbb{R}^{d_{\text{in}}}$, typically a sequence of text tokens, a UE transforms $x$ into a higher-dimensional intermediate tensor $y\in \mathbb{R}^{d_{\text{mid}}}$, that is,
\begin{equation}
y = L_U(x; \theta_{\text{UE}}),
\label{eq:edge_processing}
\end{equation}
where $\theta_{\text{UE}}$ denotes the parameters of layers in $L_U(\cdot)$. This procedure inevitably incurs a certain computational load on UEs, typically measured in floating-point operations per second (FLOPs).~Typically, for the layer $1$ with an input sequence of length $d_{\text{in}}$ and hidden dimension $d_{\text{mid}}$, utilizing a multi-head self-attention mechanism with $\kappa$ heads, the computations for the involved multi-head attention mechanism and feed-forward network components can be obtained as 
$\text{FLOPs}(L_1) = \frac{3 d_{\text{in}} d_{\text{mid}}^2}{\kappa} + \frac{2 d_{\text{in}}^2 d_{\text{mid}}}{\kappa} + 9 d_{\text{in}} d_{\text{mid}}^2$. 
Thus, the computational load on the UE can be approximated as 
\begin{align}
C_{\text{UE}}(p) & = \sum_{i=1}^{p} \text{FLOPs}(L_i) \notag \\ 
& = p \cdot \big(\frac{3 d_{\text{in}} d_{\text{mid}}^2}{\kappa} + \frac{2 d_{\text{in}}^2 d_{\text{mid}}}{\kappa} + 9 d_{\text{in}} d_{\text{mid}}^2 \big).
\label{eq:total_edge_load}
\end{align}
The intermediate tensor $y$ is transmitted from the UE to the edge over a wireless communication channel.
Mathematically, the received signal $\hat{y}\in \mathbb{R}^{d_{\text{mid}}}$ after a  Nakagami-$m$ fading channel \cite{NAKAGAMI19603, 1412063} can be represented as
\begin{equation}
\hat{y} = h \cdot y + n,
\label{eq:Nakagami_m_fading}
\end{equation}
where $h \sim \text{Nakagami}(m, \Omega)$ represents the Nakagami-$m$ fading with shape parameter $m$ and spread parameter $\Omega$, and $n \sim \mathcal{N}(0, \sigma^2)$ represents the noise following a normal distribution with zero mean and variance $\sigma^2$. In our simulation, each element of the intermediate tensor is independently subjected to a packet loss probability, abstracting each element as a separate packet to capture the impact of noise at a granular level. When $h=1$, it degenerates to an additive white Gaussian noise (AWGN) channel. 
The probability density function (PDF) of the Nakagami-$m$ distribution for the channel gain $h$ is given by:
\begin{equation}
f(h) = \frac{2m^m h^{2m-1}}{\Gamma(m)\Omega^m} e^{-\frac{m h^2}{\Omega}},
\label{eq:nakagami_pdf}
\end{equation}
where $\Gamma(m)$ denotes the Gamma function. The shape parameter $m$ controls the severity of fading. When $m = 1$, the Nakagami-$m$ distribution degenerates to a Rayleigh fading channel; at lower values of $m$, the channel experiences more severe fading, leading to greater variability in noise levels. 
In practical scenarios, particularly when splitting LLMs between UE and base stations, the mobile nature of devices often leads to rapidly changing channel conditions. For instance, as a user moves from an open outdoor area into a building or dense urban environment, the shape parameter $m$ in the Nakagami-$m$ distribution would decrease, reflecting more severe multipath fading and greater noise intensity fluctuations. This dynamic environment necessitates continuous adaptation of the splitting strategy to maintain inference performance under varying noise conditions.

~Besides, when the channel gain $h$ falls below a certain threshold $h_{\text{th}}$, the lower signal-to-noise ratio (SNR) and relatively higher bit error rate (BER) imply retransmission, thus exceeding the latency requirement in quality of service (QoS) with a rather high probability. Hence, such a case can be regarded as a packet loss. Accordingly, recalling the formula of Nakagami-$m$ distribution, the probability of packet loss can be expressed as 
\begin{align}
P(\text{Packet Loss}) &= P(h < h_{\text{th}}) \notag \\
&= \int_0^{h_{\text{th}}} \frac{2m^m h^{2m-1}}{\Gamma(m)\Omega^m} e^{-\frac{m h^2}{\Omega}} dh.
\label{eq:packet_loss_probability_nakagami}
\end{align}
\noindent Subsequently, the edge delivers an inference output $z \in \mathbb{R}^{d_{\text{out}}}$ as 
\begin{equation}
z = L_C(\hat{y}; \theta_{\text{edge}}).
\label{eq:output}
\end{equation}

The performance of the LLMs is commonly quantified using the perplexity (PPL) metric,  a standard means in NLP to evaluate how well a probability model predicts a sample.~
Given an LLM and a sequence of $N$ tokens (i.e., $w_1, w_2, \cdots, w_N$), the PPL is defined as
\begin{equation}
\text{PPL} = \exp\left( - \frac{1}{N} \sum_{k=1}^{N} \log P_{\text{LLM}}(w_k | w_1, w_2, \ldots, w_{k-1}) \right),
\end{equation}
where $P_{\text{LLM}}(w_k | w_1, w_2, \ldots, w_{k-1})$ denotes the LLM's prediction probability from the previous $k-1$ tokens.~A lower PPL signifies superior model performance, demonstrating the model's proficiency in accurately predicting the subsequent word in a sequence. Hence, in this context, PPL can serve as a unified metric to assess the impact of channel impairments on the LLM's ability.

\subsection{Problem Formulation}
\begin{figure*}[tbp]
    \centering
    \includegraphics[width=1\textwidth]{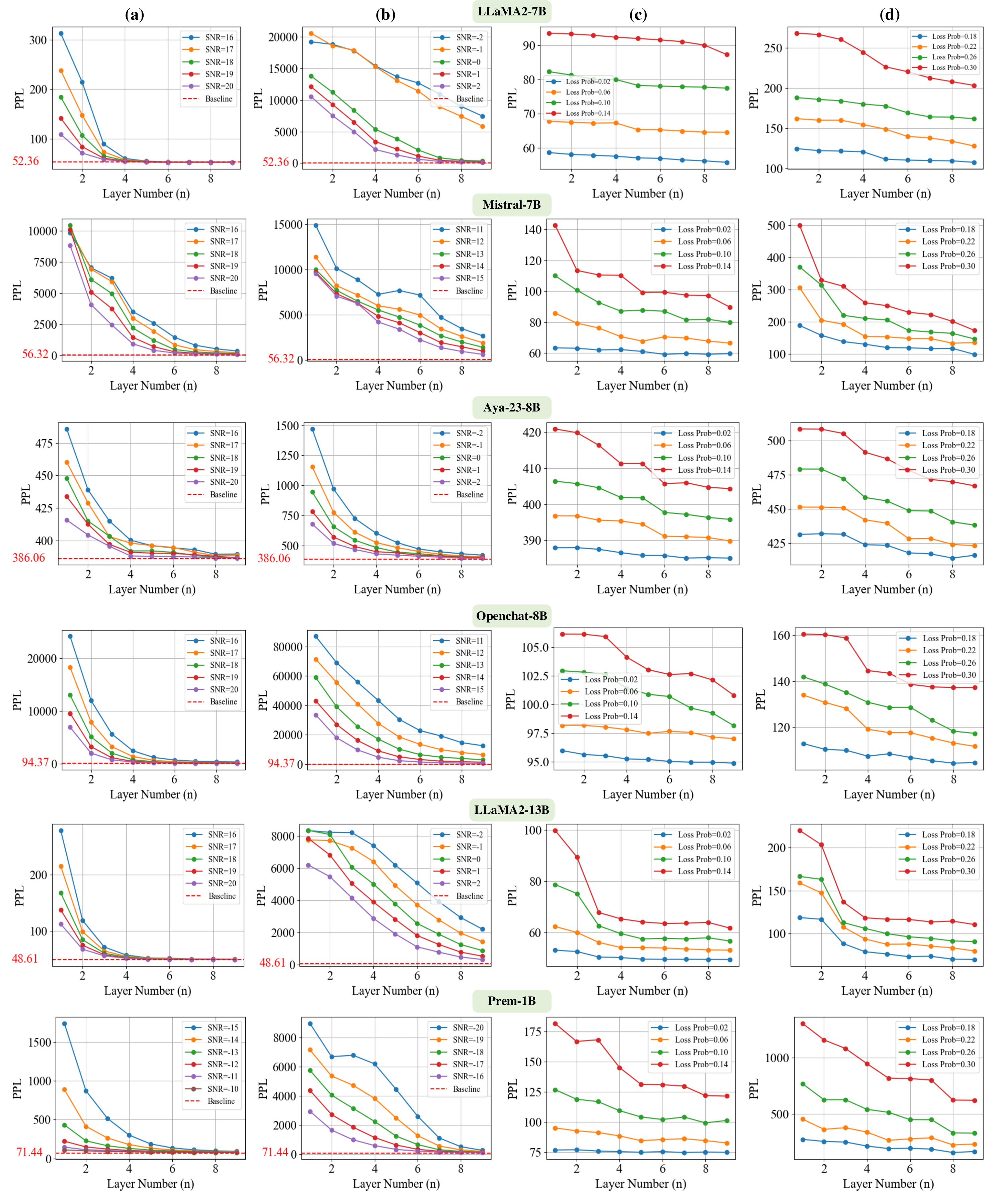}
    \caption{Illustrations of the impact on PPL across different layers for various LLMs under (a) high SNR and (b) low SNR in AWGN; (c) low packet loss probability and (d) high packet loss probability under Nakagami-$m$ fading.}
    \label{fig:noise_impact}
\end{figure*}
Beforehand, we investigate the impact of splitting points on the inference performance under different channel conditions and present the corresponding simulation results regarding several mainstream open-source LLMs, including LLaMA2-7B, LLaMA2-13B \cite{touvron2023llama}, Mistral-7B \cite{jiang2023mistral}, Aya-23-8B \cite{ustun2024aya}, Openchat-8B \cite{wang2023openchat}, and Prem-1B \cite{prem1b2024}, in Fig.~\ref{fig:noise_impact}. Consistent with our intuition, it can be observed from Fig.~\ref{fig:noise_impact} that for the same settings, a lower SNR or larger packet loss rate generally yields inferior performance (i.e., larger PPL). More interestingly, an earlier model splitting (i.e., a smaller $p$) could worsen the inference performance while the channel conditions significantly affect the performance of a given splitting point. Such an observation is also consistent with the widely recognized fact that earlier layers in LLM are responsible for learning the general or basic features of the training dataset \cite{zhang2023wider}. These findings underscore the importance of strategically selecting the splitting point in the LLM architecture to maintain desired performance.

On the other hand, high computational load on UEs would lead to increased latency and energy consumption, further undermining the benefits of deploying LLMs in edge environments.
As indicated in \eqref{eq:total_edge_load}, the computational load on the UE is approximately proportional to the number of layers processed locally.~This proportionality yields a contradicting phenomenon that an earlier model splitting worsens the inference performance but ameliorates the computational cost at the computation-limited UE. In other words, in order to minimize the overall system PPL while effectively balancing the computational load on the UE, the problem turns to identifying the optimal splitting point $p$ under volatile channel conditions, that is,
\begin{equation}
p^* = \arg\min_{p} \left[ \text{PPL}(p; \sigma, m) + \lambda \cdot C_{\text{UE}}(p) \right],
\label{eq:optimal_split}
\end{equation}
where $\text{PPL}(p; \sigma, m)$ quantifies the LLM'p inference performance, taking into account the splitting point $p$, the noise intensity $\sigma$, and the Nakagami-$m$ fading shape parameter $m$, which directly affects the packet loss probability.~Besides, the weight $\lambda$ balances the trade-off between the inference performance and computational load.

Considering the variability of network conditions and the complexities of real-time decision-making in distributed systems, we reformulate this optimization problem as a sequential decision-making task.~In that regard, RL is particularly well-suited for this scenario, due to its capability to adapt to the evolving environment and optimize decisions accordingly.

\section{Reinforcement Learning for Splitting Point Optimization}
\label{sec:rl}
In this section, we investigate the application of RL to dynamically optimize the splitting point of LLMs across UE and edge computing resources, thus adaptively responding to channel variations.

\subsection{The Markov Decision Process}

\begin{figure*}[t]
	\centering
	\includegraphics[width=0.95\textwidth]{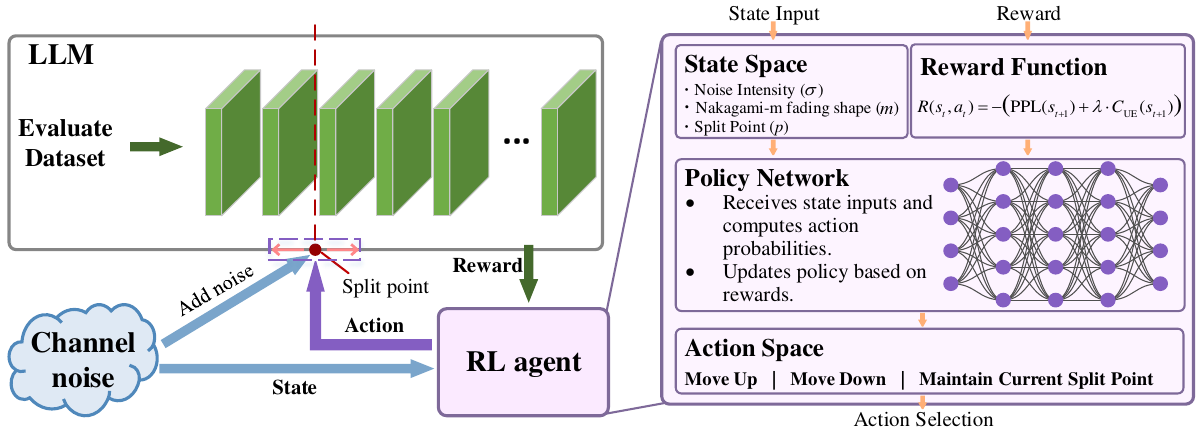}
	\caption{Illustrations of the RL setup, including the LLM, RL agent, and channel noise modules. The RL agent optimizes the splitting point of the LLM by receiving state inputs (noise intensity, Nakagami-$m$ fading shape and splitting point), computing action probabilities via the policy network, and updating the policy based on the reward function.}
	\label{fig:RL_setup}
\end{figure*}

The splitting point adjustment under volatile channels can be formalized as a Markov Decision Process (MDP), consisting of a tuple $\langle \mathcal{S},\mathcal{A}, P, R\rangle$. In particular, as illustrated in Fig. \ref{fig:RL_setup}, the state space $\mathcal{S}$ encompasses the key factors such as the noise intensity $\sigma$, the Nakagami-$m$ fading shape $m$ and the current splitting point $p$, namely, $s = \{\sigma,m,p\} \in \mathcal{S}$.
The action space $\mathcal{A}$ is designed to accommodate a range of possible adjustments to the splitting point, allowing for both fine-grained and more substantial modifications. Specifically, $\mathcal{A}$ includes actions such as moving the splitting point upward or downward by $u$ layers, where $u$ can take values such as $1$, $2$, $3$, and so forth, or maintaining the current position. Mathematically, this can be expressed as $a \in \mathcal{A} = \{-u, \dots, -1, 0, +1, \dots, +u\}$. This generalized action space provides the RL agent with the flexibility to optimize the splitting strategy dynamically in response to varying channel environments. For a time-step $t$, given an action $a_t$ under state $s_t$, the environment state will transit to the next state $s_{t+1}$ following the transition probability $P$, which is contingent on the selected action and real-time channel conditions. Meanwhile, a reward can be obtained as 
\begin{equation}
r_t = R(s_t,a_t)  = - \big( \text{PPL}(s_{t+1}) + \lambda \cdot C_{\text{UE}}(s_{t+1}) \big).
\end{equation}

Finally, the long-term overall objective can be formalized as
\begin{align}
&J(\theta) = \mathbb{E}_{\pi_\theta} \Big[ \sum_{t=1}^{T} \gamma^t \cdot r_t \Big] = \mathbb{E}_{\pi_\theta} \Big[ \sum_{t=1}^{T} \gamma^t \cdot R(s_t,a_t) \Big]\nonumber
\\
          &= \mathbb{E}_{\pi_\theta} \Big[-  \sum_{t=1}^{T} \gamma^t \cdot  \big( \text{PPL}(s_{t+1}) + \lambda \cdot C_{\text{UE}}(s_{t+1}) \big)  \Big]\label{eq:objective_function},
\end{align}
where the discount factor $\gamma$ involves the significance of future rewards. Correspondingly, it requires learning a policy $\pi_\theta$ parameterized by $\theta$ to attain the maximum of \eqref{eq:objective_function}.

\subsection{Proximal Policy Optimization}
For dynamically adjusting the splitting point of LLMs within cloud-edge-UE networks, we employ the PPO algorithm \cite{schulman2017proximal} to iteratively learn the policy $\pi_\theta$. Specifically, PPO utilizes two neural networks, namely, the policy network $\pi_\theta(a|s)$ and the value function network $V_\phi(s)$, which are parameterized by $\theta$ and $\phi$ respectively, to dictate the action $a$ given the state $s$ and estimate the expected discounted return from state $s$. Notably, PPO leverages a clipped surrogate objective function $L^{\text{CLIP}}(\theta)$, which approximates the true objective $J(\theta)$ but introduces a clipping mechanism to limit the magnitude of policy updates. Mathematically, the clipped objective can be written as

\begin{align}
L^{\text{CLIP}}(\theta) = \mathbb{E}_t \Bigg[ &\min \left( \frac{\pi_{\theta}(a_t | s_t)}{\pi_{\theta_{\text{old}}}(a_t | s_t)} \hat{A}_t, \right. \label{eq: clip_function}
\\
    &\left.\text{clip}\left( \frac{\pi_\theta(a_t | s_t)}{\pi_{\theta_{\text{old}}}(a_t | s_t)}, 1 - \epsilon, 1 + \epsilon \right) \hat{A}_t \right) \Bigg] \nonumber,
\end{align}
where 
$\theta_{\text{old}}$ represents the policy parameters for sampling.~The clipping mechanism $\text{clip}(\cdot)$ ensures that the policy ratio $\frac{\pi_{\theta}(a_t | s_t)}{\pi_{\theta_{\text{old}}}(a_t | s_t)}$ does not deviate significantly from 1, thus preventing large, destabilizing updates. This mechanism is critical in RL scenarios where stability and reliability are paramount, especially in dynamically changing environments like cloud-edge-UE networks. The advantage estimate $\hat{A}_t$, which can be computed using generalized advantage estimation (GAE), is formulated as
\begin{equation}
\hat{A}_t = \sum_{l=0}^{\infty} (\gamma \xi)^l \delta_{t+l},
\label{eq: advatage_func}
\end{equation}
where $\delta_t = r_t + \gamma V_\phi(s_{t+1}) - V_\phi(s_t)$ denotes the temporal difference error, $\gamma$ is the discount factor, and $\xi$ is the GAE parameter controlling the bias-variance tradeoff.

The update to the policy parameters $\theta$ is performed using a gradient ascent step on the clipped objective function $L^{\text{CLIP}}(\theta)$. Mathematically, 
\begin{equation}
\theta \leftarrow \theta + \alpha \nabla_\theta L^{\text{CLIP}}(\theta).
\label{eq: gradient_cal}
\end{equation}
This gradient ascent step ensures that the policy is updated iteratively to maximize the expected reward while maintaining stability through the clipping mechanism.

During the training process, we employ an experience replay mechanism specifically adapted to the dynamic nature of the channel conditions in our scenario. Notably, state-action pairs, including the current splitting point $p$, noise intensity $\sigma$, and channel fading characteristics $m$, are stored in a replay buffer. At each training step, a mini-batch of these pairs is sampled from the replay buffer to update the policy network, ensuring that the model learns from a diverse set of past experiences under varying network conditions. This technique helps to break the temporal correlations inherent in sequential channel variations, leading to more stable learning. 

\subsection{The Reward Surrogate Model for Faster RL}
\label{sec:reward}

The integration of MBRL into our LLM split optimization scenario significantly enhances the efficiency and effectiveness of our RL approach. Nevertheless, the slow reasoning capability of LLM makes the learning process sluggish. Therefore, inspired by the classical MBRL, which uses a predictive model to simulate the environment, we adopt a surrogate model to approximate the reward function, thereby boosting the learning efficiency. Specifically, we approximate the $\text{PPL}(p; \sigma, m)$ that needs to be computed by running LLM, and computes a DNN-based surrogate model $\widetilde{\text{PPL}}(p; \sigma, m,\theta_{\text{surr}})$ parameterized by $\theta_{\text{surr}}$ to minimize the 
mean squared error (MSE) as
\begin{align}
\text{MSE} = \mathbb{E} \Big[ \big( \text{PPL}(p; \sigma, m) - \widetilde{\text{PPL}}(p; \sigma, m,\theta_{\text{surr}}) \big)^2 \Big]
\label{eq: MSE_cal}.
\end{align}
Notably, we use cross-validation \cite{stone1974cross} to prevent overfitting and ensure the generalizability of the surrogate model. 

Incorporating the surrogate model directly into the reward calculation, the reward at time step $t$ can be redefined as:
\begin{align}
\tilde{R}(s_t, a_t) = -\left(\widetilde{\text{PPL}}(p_t; \sigma_t, m_t, \theta_{\text{surr}}) + \lambda \cdot C_{\text{UE}}(p_t)\right),
\end{align}
where \(\widetilde{\text{PPL}}(p_t; \sigma_t, m_t, \theta_{\text{surr}})\) represents the estimated PPL provided by the surrogate model. This reformulation significantly reduces the training burden by replacing the direct LLM inference with an efficient approximation, enabling more rapid policy evaluation and iteration of an RL policy.
From a theoretical standpoint \cite{romoff2018rewardestimationvariancereduction}, the surrogate model effectively reduces the variance in reward estimation by providing a smoothed approximation of the true reward landscape. This smoothing is particularly advantageous in high-dimensional action spaces, where small perturbations in actions could lead to large fluctuations in PPL if calculated directly.
\begin{algorithm}[!tbp]
\centering
  \caption{PPO with Reward Surrogate Model for Adaptive Splitting Point Determination in Wireless LLM Inference.}
  \label{alg:ppo-llm}
  \begin{algorithmic}[1]
  \STATE{Initialize policy network parameters $\theta$, value function network parameters $\phi$.}
  \STATE{Initialize learning rate $\alpha$, discount factor $\gamma$, GAE parameter $\xi$, clipping threshold $\epsilon$.}
  \STATE{Initialize surrogate model parameters $\theta_{\text{surr}}$.}
  \STATE{Initialize replay buffer $\mathcal{D}$.}
  \STATE{Initialize flag \textsc{use\_surrogate} $\gets \text{False}$.}
  \STATE{Initialize $\textsc{epoch\_counter} \gets 0$.}
  \STATE{Set threshold $T$ for starting surrogate model training.}
  \FOR{each training epoch}
    \STATE{$\textsc{epoch\_counter} \gets \textsc{epoch\_counter} + 1$}
    \FOR{each interaction step $t$}
        \STATE{Observe state $s_t$.}
        \STATE{Select action $a_t$ according to $\pi_{\theta}(\cdot \vert s_t)$.}
        \STATE{Execute action $a_t$ and obtain reward $r_t$. The environment transits to state $s_{t+1}$.}
        \IF{\textsc{use\_surrogate}}
            \STATE{Store transition $(s_t, a_t, s_{t+1})$ in replay buffer $\mathcal{D}$.}
        \ELSE
            \STATE{Store transition $(s_t, a_t, r_t, s_{t+1})$ in replay buffer $\mathcal{D}$.}
        \ENDIF
    \ENDFOR
    
    \STATE{Sample a mini-batch of transitions $\Phi \sim \mathcal{D}$.}
    \FOR{each transition in mini-batch}
        \IF{\textsc{use\_surrogate}}
            \STATE{Compute surrogate reward $\widetilde{\text{PPL}}(s_{t+1})$.}
            \STATE{Update advantage estimate $\hat{A}_t$ using surrogate reward with \eqref{eq: surrogat_A}.}
        \ELSE
            \STATE{Compute advantage estimate $\hat{A}_t$ using GAE from \eqref{eq: advatage_func}.}
        \ENDIF
        \STATE{Compute clipped surrogate objective $L^{\text{CLIP}}(\theta)$ with \eqref{eq: clip_function}.}
        \STATE{Perform gradient ascent on $L^{\text{CLIP}}(\theta)$ with \eqref{eq: gradient_cal}.}
    \ENDFOR
    \IF{not \textsc{use\_surrogate} \AND \\ 
        $\textsc{epoch\_counter} \geq T$}
        \STATE{Train surrogate model $\widetilde{\text{PPL}}(p; \sigma, m,\theta_{\text{surr}})$ to minimize MSE with \eqref{eq: MSE_cal}.}
        \STATE{Set \textsc{use\_surrogate} $ \gets \text{True}$.}
    \ENDIF
  \ENDFOR
\end{algorithmic}
\end{algorithm}

Correspondingly, during the training process of RL, $\hat{A}_t$ in \eqref{eq: advatage_func} can be re-written as
\begin{align}
    \hat{A}_t &= \sum_{l=0}^{\infty} (\gamma \xi)^l \Big[ \big( \widetilde{\text{PPL}}(s_{t+1+l}) + \lambda \cdot C_{\text{UE}}(s_{t+1+l}) \big) \notag \\
    & \qquad \qquad \qquad + \gamma V(s_{t+2+l}) - V(s_{t+1+l}) \Big]. 
\label{eq: surrogat_A}
\end{align}
On this basis, we can compute $L^\text{CLIP}(\theta)$, which consequently facilitates the update of $\theta$.

By incorporating this model-based approach, as shown in Table \ref{Table:RL_MBRL_Comparison}, we achieve substantial gains in computational efficiency, enabling the RL agent to flexibly accommodate changing deployment environments.

Finally, we summarize the algorithm in Algorithm \ref{alg:ppo-llm}.

\section{Simulation Settings and Experimental Results}
\label{sec:simulation}

\begin{figure*}[t]
	\centering
	\includegraphics[width=0.95\textwidth]{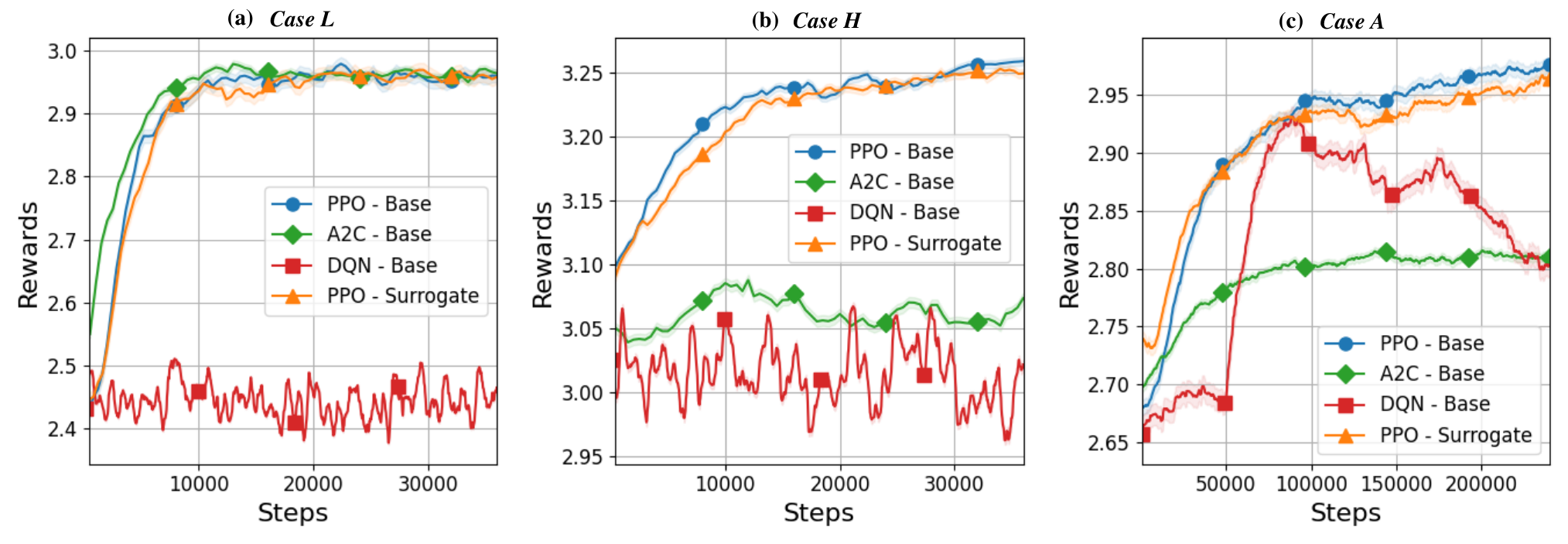}
	\caption{Comparison of training performances for different RL approaches under \emph{Case L}, \emph{Case H}, and \emph{Case A}.}
	\label{fig:reward_trend}
\end{figure*}

\subsection{Experimental Setup}

To validate the effectiveness of RL-based adaptive LLM splitting point determination, we use the LLaMA2-7B model \cite{touvron2023llama}, a 32-layer LLM, as well as the WikiText-2 dataset \cite{merity2016pointer}, which contains $4,355$ sentences with an average length of $20$ words, to evaluate the PPL under varied network conditions.~Particularly, we simulate a changing, fading-induced packet loss probability in the range between $0$ and $0.3$. Moreover, we primarily consider three representative cases: 
\begin{itemize}
\item \emph{Case L}: A low packet loss probability $0\sim0.1$ and an initial splitting point $p_{\text{init}_L}$ near the input (layers $1$-$5$).
\item \emph{Case H}: A high packet loss probability $0.1\sim0.3$ and an initial splitting point $p_{\text{init}_H}$ far from the input (layers $6$-$10$).
\item \emph{Case A}: Complete range of packet loss probability $0\sim0.3$ and initial splitting points $p_{\text{init}_A}$ (layers $1$-$10$).
\end{itemize}
Besides, the default hyperparameters for the PPO \cite{schulman2017proximal} algorithm and the channels are given in Table \ref{tab:Hyperparameters}. 

\begin{table}[t]
\centering
\caption{PPO Algorithm Hyperparameters} \label{tab:Hyperparameters}
\renewcommand{\arraystretch}{1.25}
\setlength{\tabcolsep}{5pt}
\begin{tabular}{p{4cm} p{2cm}}
    \toprule[0.75pt]
    \textbf{Hyperparameter} & \textbf{Value} \\
    \midrule[0.5pt]
    Learning rate ($\alpha$) & $0.0003$ \\
    Discount factor ($\gamma$) & $0.99$ \\
    Clipping parameter ($\epsilon$) & $0.2$ \\
    Update frequency ($n_{\text{step}}$) & $400$ \\
    Batch size & $100$ \\
    Steps per episode & $5$ \\
    GAE ($\xi$)& $0.95$ \\
    \bottomrule[0.75pt]
\end{tabular}
\end{table}

To reduce the computational cost of evaluating the LLM's performance at each step, by collecting $9,718$ pieces of practical records, we derive a reward surrogate model as in Section \ref{sec:reward}. Our result shows that an MLP (Multi-Layer Perceptron) yields a test loss of $0.00548$ in MSE and $0.050$ in mean absolute error (MAE), thus providing sufficient accuracy. Therefore, we use this MLP-based reward surrogate model to accelerate the evaluation process.

\subsection{Experiment Results}

We first present the performance of PPO with and without the reward surrogate model and compare them with baseline RL schemes (i.e., A2C \cite{Mnih2016} and DQN \cite{mnih2015human}). Fig. \ref{fig:reward_trend} presents the corresponding results.~It can be observed from Fig.~\ref{fig:reward_trend} that for \emph{Case H} and \emph{Case A}, PPO yields significantly superior performance than A2C and DQN; while for \emph{Case L}, all RL approaches lead to similar performance. Besides, PPO trained with reward surrogate models closely resemble that with actual rewards. Furthermore, Table \ref{Table:RL_MBRL_Comparison} compares PPO with and without reward surrogate model under \emph{Case A} in terms of reward, training duration, and computational resource consumption. The results indicate that the reward surrogate model significantly reduces the training time and computational resource consumption while achieving comparable rewards.

\begin{table*}[thb]
\centering
\caption{Performance comparison of PPO with and without reward surrogate model under \emph{Case A}.} \label{Table:RL_MBRL_Comparison}
\setlength{\tabcolsep}{2pt} 
\begin{tabular}{lcc}
    \toprule[0.75pt]
    \textbf{Metric} & \textbf{w.o. surrogate} & \textbf{w. surrogate} \\
    \midrule[0.5pt]
    Reward at $24,000$ Steps & $2.9736$ & $2.9663$ \\
    Training Duration & $> 24$ days & $7.7$  minutes \\
    Computational Resource Consumption & $16.3$ GB & $< 1$ GB \\
    \bottomrule[0.75pt]
\end{tabular}
\end{table*}

\begin{figure}[!t]
    \centering
    \includegraphics[width=0.95\columnwidth]{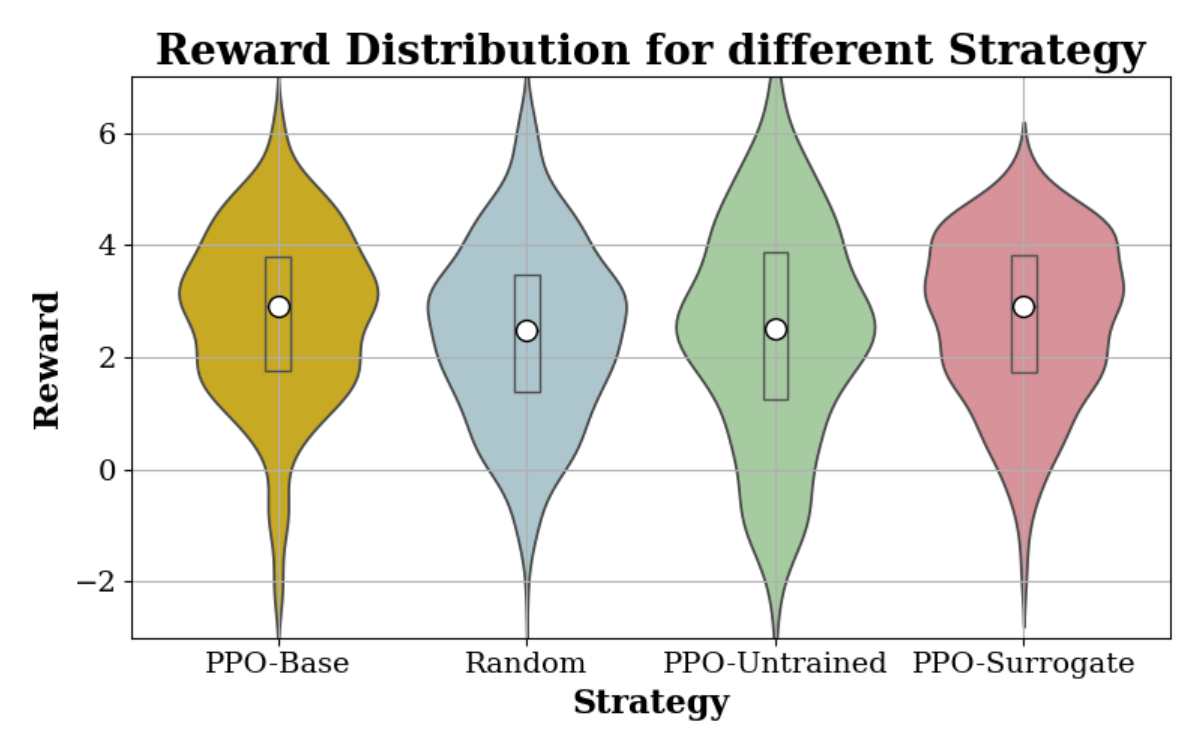}
    \caption{This violin plot compares the reward distributions across four different strategies. The width of each violin represents the density of rewards at different values, with wider sections indicating a higher probability of observing rewards in that range. The central white dot represents the median reward, while the thick black bar in the center denotes the interquartile range (IQR). }
    \label{fig:reward_distributed}
\end{figure}
\begin{figure}[!t]
	\centering
	\includegraphics[width=0.475\textwidth]{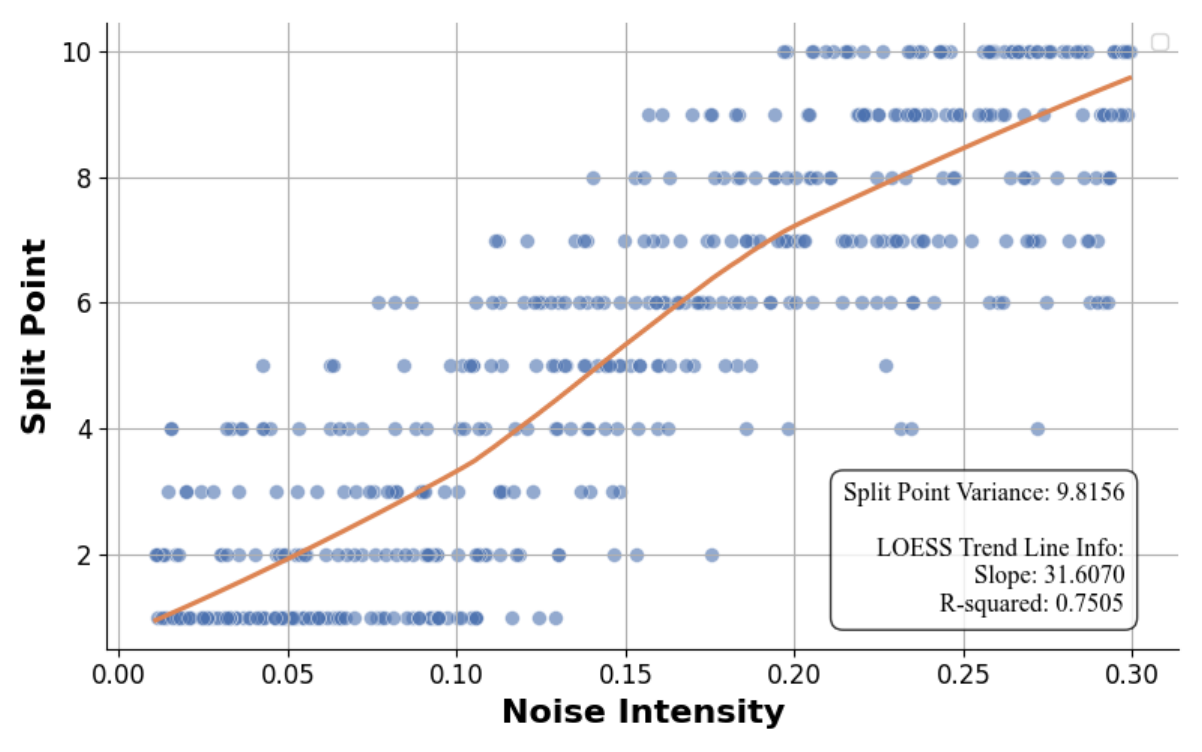}
	\caption{splitting points determined by the trained PPO agent across different noise intensities, along with a LOESS trend line.}
	\label{fig:cloud_point}
\end{figure}
\begin{figure}[t]
	\centering
	\includegraphics[width=0.475\textwidth]{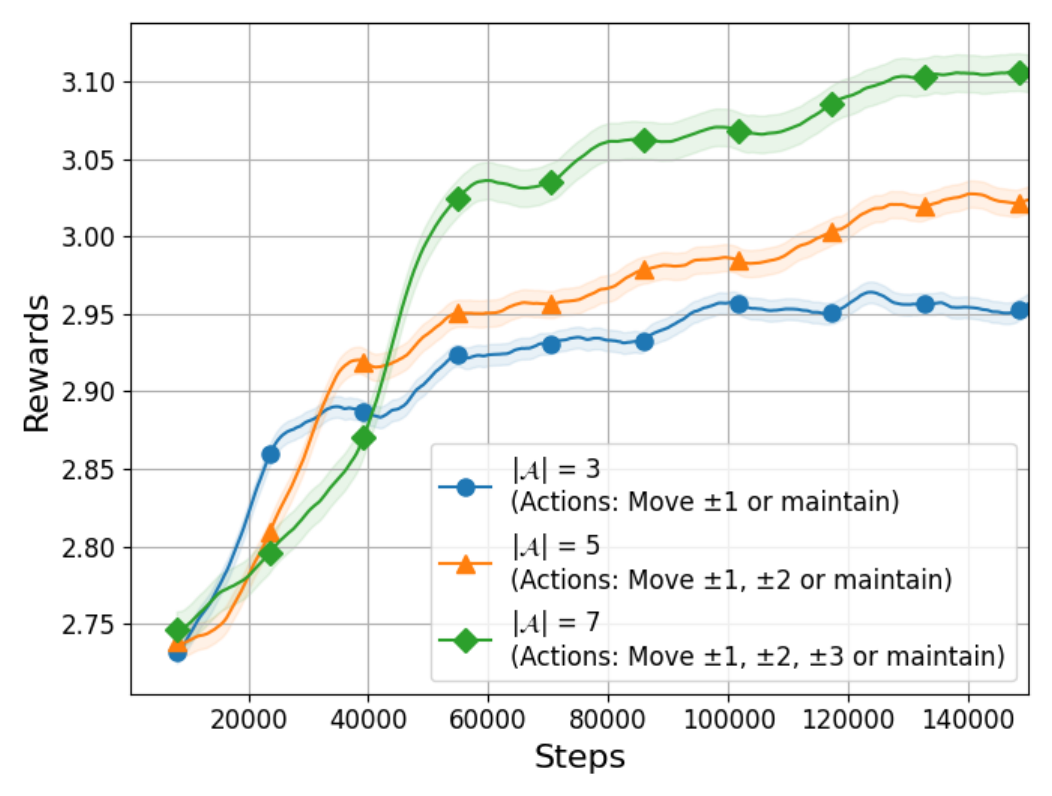}
	\caption{Comparison of training performances across different action spaces (Action Space = 3, 5, 7) for varying movement ranges.}
	\label{fig:action_space_contrast}
\end{figure}
Fig.~\ref{fig:reward_distributed} presents a violin plot comparing the reward distributions of four different strategies: the trained PPO agent with true reward training, the trained PPO agent using the reward surrogate model, a random policy, and an untrained PPO agent.
The plot shows that the trained agents, both standard and MBRL-enhanced, lead to higher average rewards and tighter reward distribution, indicating more consistent and superior performance compared to the random and untrained agents.

\begin{figure*}[!h]
    \centering
    \includegraphics[width=0.85\textwidth]{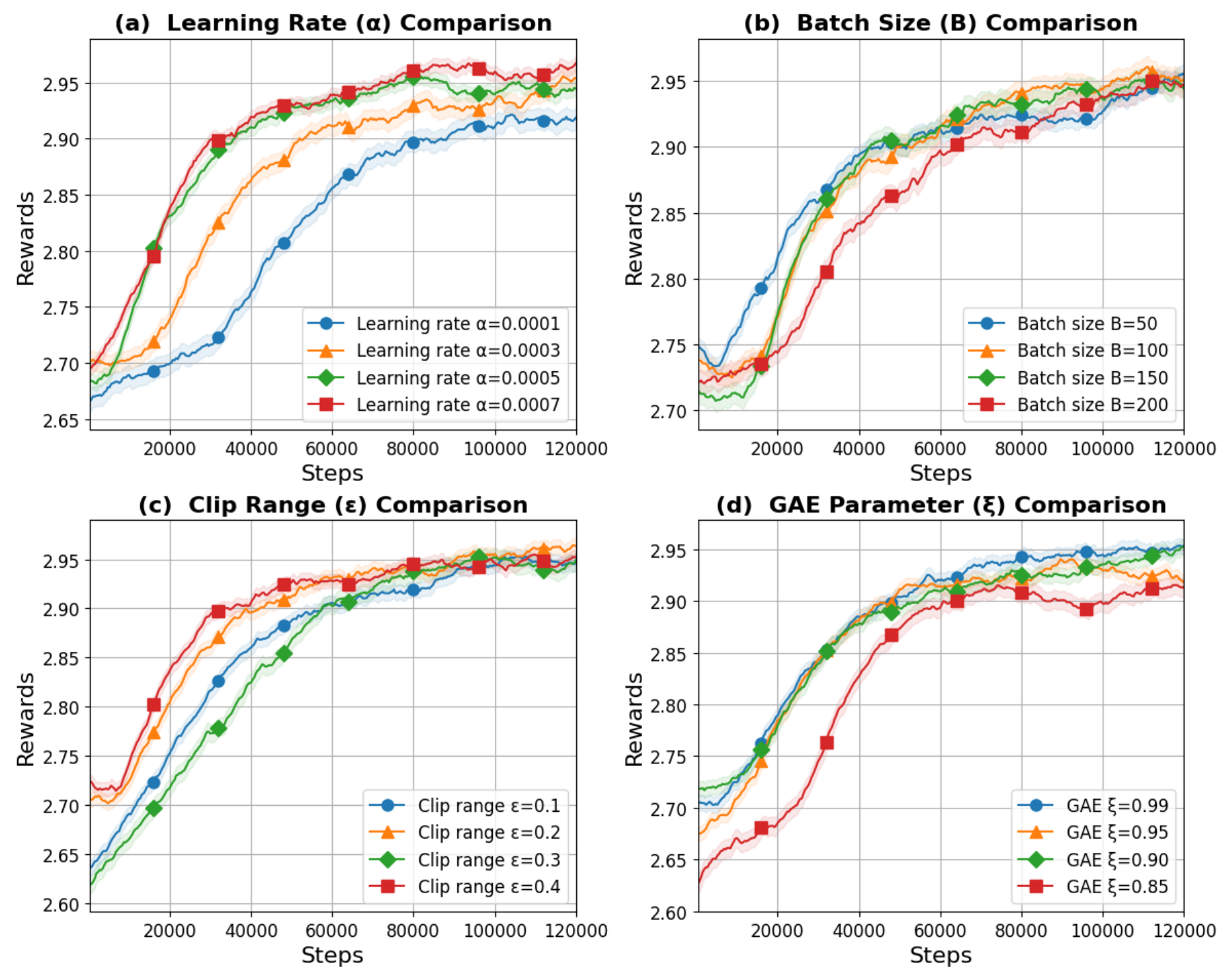}
    \caption{Impact of various hyperparameters with respect to training steps in the PPO algorithm, including comparisons of learning rate $\alpha$, batch size $B$, clip range $\epsilon$, and GAE parameter $\xi$.}
    \label{fig:reward_trend_contrast}
\end{figure*}

Fig. \ref{fig:cloud_point} illustrates $500$ splitting points determined by the trained PPO agent under \emph{Case A}, and complements a locally estimated scatterplot smoothing (LOESS) \cite{Cleveland1979} trend line, whose slope and R-squared value provide quantitative insights into the relationship between channel conditions and splitting point decisions. It can be observed from Fig. \ref{fig:cloud_point} that as noise intensity $\sigma$ increases, the agent prefers to place the splitting point further from the input layers. This strategic adjustment helps mitigate the adverse effects of noise on model performance by leveraging the cloud's more robust processing capabilities.

In addition to the baseline experiment where the action space is limited to single-layer adjustments, we conduct further experiments with enlarged action space to evaluate the impact of larger adjustments on the training process and final rewards. As shown in Fig. \ref{fig:action_space_contrast}, allowing larger adjustments (e.g., moving by $±2$, $±3$ layers) leads to slower convergence but ultimately achieves higher rewards. This trade-off suggests that while the larger action space can explore a wider range of configurations, they may require more training steps to stabilize. However, in both experimental and practical scenarios, single-layer adjustments offer notable advantages. They provide fine-grained control over the splitting point, allowing the model to quickly adapt to changes in channel conditions. This is particularly beneficial in dynamic environments where frequent and subtle adjustments are necessary to maintain optimal performance.

The impact of various hyperparameter settings on PPO training performance is analyzed in Fig.~\ref{fig:reward_trend_contrast}. 
Fig.~\ref{fig:reward_trend_contrast}(a) indicates that higher learning rates ($\alpha=0.0005$ and $\alpha=0.0007$) lead to faster initial learning but may introduce higher variance in the rewards. Fig.~\ref{fig:reward_trend_contrast}(b) demonstrates that larger batch sizes ($B=150$ and $B=200$) generally result in smoother and more stable reward curves, yielding better gradient estimates. Fig.~\ref{fig:reward_trend_contrast}(c) reveals that moderate clip ranges ($\epsilon$=0.2 and $\epsilon$=0.3) strike a balance between stability and performance, whereas too small or too large clip ranges can degrade performance. Fig.~\ref{fig:reward_trend_contrast}(d) presents that a larger GAE parameter ($\xi$=0.99) produces more impressive long-term reward accumulation, emphasizing the importance of temporal smoothing in advantage estimation.

\section{Conclusion and Future Works}
\label{sec:conclusion}
In this paper, we have presented an MBRL framework for dynamically optimizing the splitting point of LLMs deployed across UE and the edge, so as to enhance the efficiency and performance of LLMs under wireless network conditions. In particular, we have formulated the problem as an MDP, and introduced a reward surrogate model to significantly shorten overall training time. 
The experimental results have demonstrated the framework's efficacy in managing the trade-off between inference performance and computational load at the UE. Meanwhile, comprehensive validations in mainstream open-source LLMs have clearly demonstrated that an earlier model splitting could worsen the point inference performance, which might provide an independent interest to the community.
Our proposed framework offers a structured approach to dynamically deploying LLMs across heterogeneous devices. In practical applications, such as in smart cities and industrial IoT, this framework can enhance the flexibility of LLM deployment while alleviating the computational constraints associated with running LLMs on edge devices.

Despite these achievements, several limitations and challenges remain.~Though the validation of the impact of splitting points on the performance of some widely adopted LLMs, given the versatility of LLMs, the generality issue still awakens further attention.  The lack of a more accurate channel model and the absence of communication-efficient distribution learning approaches (e.g., quantization) in data transmission also demands future research.
Also, future research could explore adaptive mechanisms that dynamically adjust the action space based on the current learning phase or environmental conditions, thereby balancing the need for quick convergence and high reward attainment.
Additionally, further investigation is required into the scalability of our framework in larger, more complex network environments and its generalization across different LLM architectures.
We will explore these important directions in the future. 



\end{document}